\documentclass{amos}

\usepackage{amsmath}
\usepackage{amssymb}
\usepackage{dsfont} 
\usepackage{xcolor}
\usepackage{hyperref}
\usepackage{cleveref}
\usepackage{siunitx}
\DeclareMathOperator*{\argmax}{arg\,max}
\DeclareMathOperator*{\argmin}{arg\,min}
\DeclareMathAlphabet{\mathcal}{OMS}{cmsy}{m}{n} % force latex font for math writing / equations

\usepackage{booktabs}

\title{\TitleFont Chance-Constrained Belief-Space Maneuver Planning \\ for Autonomous Collision Avoidance Under Uncertainty}

\author{Grace Ra Kim \\ Stanford University \and Duncan Eddy and Mykel J. Kochenderfer \\ Stanford University}

\date{}

\begin{document} 

\maketitle

\begin{abstract}
	\normalsize
	Increasing conjunction frequency in low Earth orbit places growing pressure on spacecraft operators to determine not only whether an encounter requires mitigation, but whether sufficient information is available to commit to a maneuver. Delaying action may provide additional tracking information that reduces state uncertainty and resolves an apparent conjunction. However, waiting also reduces the time and maneuver authority available if intervention ultimately becomes necessary, potentially increasing the required propellant expenditure. This work formulates this information--action tradeoff as a belief-space planning problem for conjunctions between a maneuverable spacecraft and an unmaneuverable secondary object. The planner represents the uncertain orbital states as Gaussian beliefs and uses a chance-constrained belief-space Monte Carlo tree search framework to reason over possible future tracking updates before time of closest approach (TCA). Because these observations are not known in advance, the resulting belief at TCA is uncertain, as different possible observation sequences may produce different terminal belief states. A terminal chance constraint limits the probability of reaching TCA with a belief whose probability of collision exceeds a prescribed threshold, allowing the planner to wait when future tracking is likely to resolve the conjunction while requiring intervention when continued deferral becomes too risky. We evaluate the approach on eight historical conjunctions from NASA's Conjunction Assessment Risk Analysis dataset. By varying the secondary-object measurement quality and tracking cadence, we generate a total of 96 distinct evaluation scenarios. Across the evaluated conditions, the planner reaches TCA without maneuvering in approximately 40\% of episodes while maintaining no terminal collision-risk violations. In contrast, fixed-time rule-based maneuver policies resolve more encounters without maneuvering when intervention is deferred closer to TCA, but at the expense of increasing terminal risk violations. The ability to delay maneuver decisions depends strongly on tracking quality and measurement cadence, ranging from 76\% under accurate, frequent measurements to approximately 18\%–20\% under the poorest tracking conditions. For the same physical conjunction, changing only the expected tracking quality can shift the preferred decision from continued waiting to delayed maneuver commitment or immediate intervention. These results show that tracking quality and frequency are not only inputs to collision-risk estimation: they can determine when intervention becomes necessary. Explicitly valuing future information therefore provides a principled mechanism for deciding when an avoidance decision can safely remain open and when uncertainty itself justifies action.
\end{abstract}
\section{Introduction}
\label{sec:introduction}

Low Earth orbit (LEO) is becoming increasingly congested, increasing the demands placed on the systems and operators responsible for conjunction assessment. In 2023, the United States Space Force 18th Space Defense Squadron (18 SDS) generated approximately 600,000 Conjunction Data Messages (CDMs) per day, three times the average daily rate reported in 2020~\cite{ramos2023lessons}. Behind this aggregate volume, however, each conjunction unfolds as a relatively short sequence of evolving risk assessments. As new tracking data become available, state estimates and covariances are refined, collision-risk estimates are updated, and operators must continually reassess whether a maneuver is warranted. At the same time, the information available for any one encounter can be sparse. NASA's Conjunction Assessment Risk Analysis (CARA) program reports that conjunctions are generally first identified within seven days of the time of closest approach (TCA), yet most have fewer than ten CDMs available over that period~\cite{mashiku2025cara}. Collision avoidance therefore presents a sequential decision problem in which operators must determine not only whether an encounter poses sufficient risk to warrant action, but also when the available information is sufficient to commit to a maneuver.

This timing question is particularly important for conjunctions between a maneuverable spacecraft and an unmaneuverable secondary object. Although coordination between active operators has become increasingly important in dense lower-LEO constellation shells~\cite{kim2026semidecentralized}, ESA reports that conjunctions at higher LEO altitudes remain dominated by debris~\cite{esa2026ser10}. More generally, an ``active'' classification does not necessarily imply maneuverability, as ESA classifies an object as active based on the existence of a communication link and explicitly notes that such an object may nevertheless be nonmaneuverable. When the secondary cannot coordinate an avoidance action or provide an authoritative maneuver plan, the primary spacecraft bears the burden of conjunction mitigation, while knowledge of the secondary is expected to improve through future surveillance observations. In this setting, choosing to wait is a deliberate action. Delaying a maneuver allows the planner to incorporate additional tracking information before committing to an avoidance action, but simultaneously reduces the remaining time and fuel efficiency and mitigation effectiveness available if intervention ultimately becomes necessary.

Operational practice already reflects this tradeoff. Conjunction decisions are commonly informed by the probability of collision, $P_c$, evaluated against prescribed action criteria. For example, NASA specifies a baseline mitigation threshold of $P_c>10^{-4}$ at the mitigation commitment point~\cite{nasa2023npr8079}. Operators also differ in maneuver timing, with some favoring early intervention and others delaying action to refine the estimated encounter geometry and collision risk~\cite{alfano2022operators}. Delaying intervention may also avoid the operational impact of an unnecessary maneuver when an uncertain conjunction later resolves. Existing collision avoidance research has developed methods for collision-risk assessment and fuel-efficient maneuver design under state uncertainty~\cite{akella2000probability,alfriend1999probability,bombardelli2015}. More recently, chance constraints have been incorporated directly into spacecraft collision-avoidance optimization. Ryu et al.\ formulate fuel-optimal debris avoidance with a distributionally robust chance constraint on collision probability~\cite{ryu2025risk}, while Arzelier et al.\ use sample-average approximation to minimize maneuver cost subject to a prescribed collision-probability threshold~\cite{arzelier2025saa}. Chance-constrained control has also been applied to spacecraft trajectory planning under navigation, maneuver-execution, and dynamical uncertainty~\cite{oguri2024chance}. These approaches demonstrate how collision risk can be treated explicitly as a probabilistic constraint rather than only as a term in the maneuver objective. In these formulations, however, uncertainty primarily enters the optimization of the maneuver or trajectory itself. Future observations are not treated as part of the decision process, and minimal explicit reasoning is performed about the information that could be gained at future stages.

A separate line of recent work has begun to treat collision avoidance as a sequential decision problem. Bourriez et al. formulate autonomous collision avoidance as a POMDP with imperfect debris monitoring~\cite{bourriez2023spacecraft}. Kim et al. extend this decision-theoretic perspective to multi-spacecraft collision avoidance, using a semi-decentralized POMDP to account for intermittent information sharing under realistic communication constraints~\cite{kim2026semidecentralized}. Kuhl et al. consider online maneuver planning against a nonmaneuverable object using an MDP~\cite{kuhl2025mdp} and Ferrara et al. explicitly examine the trade between delaying a maneuver for improved conjunction information and preserving the opportunity for effective intervention~\cite{ferrara2025mdp}. These works establish that maneuver timing and evolving information belong within the decision process itself. They do not, however, impose the terminal safety requirement considered in this work: a bound on the probability, over possible future observations and resulting belief evolution, that the collision-risk criterion will be violated at TCA.

While these prior works motivate a sequential decision formulation of collision avoidance, they do not explicitly constrain risk over the possible belief states induced by future tracking observations. Because those observations are not known in advance, different measurement sequences may produce different beliefs at TCA, potentially resolving the conjunction or reinforcing the need to maneuver. The relevant safety question is therefore not only whether the current probability of collision satisfies a prescribed threshold, but how likely future observations are to produce a terminal belief that violates that safety threshold.

In this work, we formulate autonomous collision avoidance as a finite-horizon, belief-space planning problem for a maneuverable primary spacecraft and an unmaneuverable secondary. The planner maintains Gaussian beliefs over both orbital states and uses a chance-constrained Monte Carlo tree search (MCTS)~\cite{browne2012mcts,moss2024constrainedzero} planner to reason over possible future tracking observations before TCA. Because these observations are not yet known, each candidate action induces a distribution over possible terminal beliefs. We impose a chance constraint on this distribution, requiring the probability of reaching TCA with a belief whose collision probability exceeds a prescribed threshold to remain below a specified risk level. This construction separates the collision-probability threshold that defines an unsafe terminal belief from the likelihood that future observations lead to such a belief. By applying the constraint at TCA, the planner can intelligently continue to wait when additional tracking is likely to resolve the conjunction and commit to a maneuver when continued deferral would no longer satisfy the prescribed terminal risk requirement.

There are three main contributions of this work. First, we formulate the maneuverable--non-maneuverable maneuver decision as a terminal chance-constrained belief-space planning problem, allowing future tracking information to influence the decision to wait or maneuver before that information is observed. Second, we evaluate the resulting planner on real NASA CARA conjunction scenarios and compare its decisions against rule-based and optimistic greedy offline-planning policies, characterizing the trade between safely deferring a maneuver or committing to a maneuver early. Third, we examine how this behavior changes with measurement accuracy and tracking cadence, identifying the transition between tracking regimes in which uncertainty materially changes the maneuver decision. Collectively, these results establish chance-constrained belief-space planning as a principled framework for autonomous collision avoidance that explicitly accounts for the decision value of future tracking information.

\section{Problem Formulation}
\label{sec:problem_formulation}

We formulate spacecraft collision avoidance as a finite-horizon sequential decision problem under evolving state-estimate uncertainty. The central decision is whether an avoidance maneuver is necessary and, if so, whether to execute it immediately or wait for additional tracking that may resolve the conjunction without intervention. Acting earlier provides more time for a maneuver to modify the encounter geometry, but requires committing before additional tracking can determine whether intervention is necessary. Waiting allows future observations to refine the estimated state and uncertainty, but reduces the remaining time available for intervention. We represent this information--action tradeoff as a belief-space decision problem and impose the probability of collision $P_c$ as the probabilistic safety quantity.

\subsection{Conjunction Scenario and Scope}
\label{subsec:scenario_scope}

We consider a conjunction between one maneuverable active spacecraft, denoted by $s$, and one nonmaneuvering secondary resident space object, denoted by $d$. The secondary may represent orbital debris, a rocket body, or another resident space object for which no avoidance action is modeled. The Cartesian states of the primary and secondary objects in the Earth-centered inertial (ECI) frame are
\begin{equation}
x^s =
\begin{bmatrix}
r_x^s &
r_y^s &
r_z^s &
v_x^s &
v_y^s &
v_z^s
\end{bmatrix}^{\top}
\in \mathbb{R}^{6},
\qquad
x^d =
\begin{bmatrix}
r_x^d &
r_y^d &
r_z^d &
v_x^d &
v_y^d &
v_z^d
\end{bmatrix}^{\top}
\in \mathbb{R}^{6}
\end{equation}
where the first three components are position and the final three components are velocity. The planning horizon for a conjunction event starts with the generation of an initial CDM for the conjunction and terminates at the time of closest approach. We denote the time remaining to TCA by $\tau$, such that $\tau=0$ corresponds to the TCA.

This work considers conjunctions for which a two-dimensional encounter-plane probability of collision provides an appropriate representation of risk. In particular, we consider short-duration encounters for which the relative motion near TCA is sufficiently well represented by the assumptions underlying the 2D collision-probability model. Encounters outside this validity regime were not considered in the present experiments. The planning framework itself is not restricted to this regime, but extension to these encounters would require an appropriate alternative risk metric, such as a higher-fidelity $P_c$ method, which may come at greater computational cost. %The criteria used to characterize this regime and the validation of the evaluation conjunction set against these criteria is discussed in~\Cref{sec:experimental_setup}.

%Within this scope, the goal is to determine how a maneuver decision should evolve as additional information becomes available before TCA. We therefore separate the maneuver-timing problem considered here from the more general problem of jointly optimizing maneuver direction and magnitude.
\subsection{Belief-MDP Formulation}
\label{subsec:belief_mdp}

Sequential collision avoidance requires a planner to repeatedly decide to maneuver or wait as the estimated encounter advances towards TCA. If the physical state of the conjuncting objects were known exactly, this problem could be represented as a Markov decision process (MDP), in which each action is selected from the current state to optimize a long-term objective~\cite{puterman1994}. In practice, however, the primary spacecraft and secondary object orbital states are not known exactly. They are inferred from imperfect tracking measurements and are therefore uncertain.

This uncertainty makes the underlying problem partially observable. A partially observable Markov decision process (POMDP) extends the MDP framework by allowing the planner to receive noisy or incomplete observations of an underlying physical state rather than observing that state directly~\cite{kaelbling1998,kochenderfer2022algorithms}. The planner must therefore base its decisions on the belief, or the distribution of potential states.

A POMDP can be reformulated as an equivalent MDP over belief-states, where each belief is a probability distribution over the possible underlying physical states~\cite{kochenderfer2022algorithms,smallwood1973}. This representation is particularly useful for collision avoidance because $P_c$ depends on both the estimated encounter geometry and its associated uncertainty. As new observations are received, the belief is updated, allowing the planner to recompute $P_c$ and reason directly over the evolving state uncertainty. We therefore formulate collision avoidance as a finite-horizon belief-MDP $\mathcal{M}_b$, with the current belief $b$ serving as the state for both risk evaluation and maneuver planning
\begin{equation}
\mathcal{M}_b
=
\left(
\mathcal{B},
\mathcal{A},
T,
\mathcal{R},
%\gamma,
\mathcal{H}
\right)
\end{equation}
where $\mathcal{B}$ is the belief space, $\mathcal{A}$ is the action space, $T$ is the transition model over beliefs, $\mathcal{R}$ defines the reward or decision objective, 
%$\gamma$ is the discount factor, 
and $\mathcal{H}$ is the planning horizon. Decisions are made at a finite set of prescribed times before TCA. We index these decision epochs by $k$, with monotonically decreasing $\tau_k$ denoting the time remaining to TCA at epoch $k$. At each epoch, the planner maintains a belief $b_k$ representing its current probabilistic estimate of the conjunction state. Because future observations are not known when an action is selected, the same current belief and action may lead to different successor beliefs. The transition model $T$ therefore represents the distribution over possible successor beliefs induced by the future observation. Once a particular observation is realized, the corresponding posterior belief is obtained through a belief update model. %Future observations are sampled explicitly during planning and used to generate possible successor beliefs, allowing the planner to reason over how the conjunction estimate may evolve before TCA.
\subsubsection{Belief Space}
\label{subsubsec:belief_space}
We represent the state uncertainty of the primary and secondary objects at decision epoch $k$ by multivariate Gaussian beliefs, parameterized by their respective state means and covariances:
\begin{equation}
x_k^s
\sim
\mathcal{N}
\Big(
\mu_k^s,
\Sigma_k^s
\Big),
\qquad
x_k^d
\sim
\mathcal{N}
\Big(
\mu_k^d,
\Sigma_k^d
\Big)
\end{equation}
where $\mu_k^s,\mu_k^d\in\mathbb{R}^{6}$ are the estimated Cartesian states and $\Sigma_k^s,\Sigma_k^d\in\mathbb{R}^{6\times6}$ are the corresponding state covariance matrices. The belief $b_k \in \mathcal{B}$ used for planning is therefore represented by
\begin{equation}
b_k =
\left(
\mu_k^s,
\Sigma_k^s,
\mu_k^d,
\Sigma_k^d,
\tau_k
\right)
\end{equation}
which contains the estimated state and uncertainty of both objects together with the time remaining to TCA. %The primary and secondary are represented by independent six-dimensional Gaussian beliefs rather than a single coupled twelve-dimensional distribution. 
We assume independent state estimation errors between the two objects, eliminating the need to maintain cross-correlation terms and reducing the dimensionality of the belief representation.
\subsubsection{Action Space}
\label{subsubsec:action_space}

At each decision epoch, the planner selects an action $a_k \in \mathcal{A} = \left\{\text{WAIT},\text{MANEUVER}\right\}$, corresponding to either waiting for additional information or executing an avoidance maneuver. Here, $\text{WAIT}$ applies no velocity change and $\text{MANEUVER}$ applies a fixed impulsive maneuver to the active spacecraft. The maneuver magnitude is fixed at $\Delta v^{\mathrm{man}} = 0.1$~m/s, a conservative but operationally plausible magnitude for a low-Earth-orbit collision-avoidance maneuver~\cite{bombardelli2015,klinkrad2005}, and is applied in the prograde direction according to
\begin{equation}
\Delta \mathbf{v}_s
=
\Delta v^{\mathrm{man}}
\frac{\mathbf{v}_s}
{\lVert \mathbf{v}_s\rVert}
\end{equation}
When the planner selects the \text{MANEUVER} action, this impulsive velocity change is applied immediately at the current decision epoch. The planner therefore determines, at each decision epoch, whether to execute the fixed maneuver or continue waiting, thereby selecting the maneuver timing from the discrete set of available decision epochs. Holding the maneuver magnitude and direction fixed isolates the effect of maneuver timing across the discrete decision epochs under evolving state uncertainty.

\subsubsection{Transition and Observation Model}
\label{subsubsec:transition_observation}

The belief evolves from one decision epoch to the next according to the selected action and the tracking information received before the next decision. Given the current belief $b_k$ and action $a_k$, the belief transition model $T(b_{k+1}\mid b_k,a_k)$ describes the distribution over possible successor beliefs $b_{k+1}$ at the next decision epoch. These transitions occur over the finite set of prescribed times $\mathcal{E}=\{\tau_0,\tau_1,\ldots,\tau_{\text{TCA}}=0\}$, where $\tau_k$ denotes the time remaining to TCA at epoch $k$. Intermediate decision epochs occur at a prescribed fixed cadence corresponding to scheduled secondary tracking updates. At each decision epoch, the secondary belief is updated using the newly received measurement, the primary belief is updated using its assumed onboard navigation information, and the planner selects WAIT or MANEUVER. %The terminal epoch $\tau_{\text{TCA}}=0$ corresponds to TCA.

Future observations are not known when $a_k$ is selected, so the same current belief and action may lead to different successor beliefs depending on the observation received before the next decision epoch. At each scheduled tracking opportunity, the current Gaussian belief is propagated under action $a_k$ to the next decision epoch and updated using the received observation $z_{k+1}$ through a Kalman measurement update. The propagation and measurement-update models are detailed in~\Cref{subsec:belief_tracking}. During planning, future observations are sampled from the observation model, producing different possible successor beliefs and allowing the planner to reason over how future tracking may change the estimated encounter geometry and uncertainty before TCA.

% Future observations are not known when $a_k$ is selected, so the same current belief and action may lead to different successor beliefs depending on the observation received before the next decision epoch. At a scheduled tracking opportunity, the planner receives an observation $z_{k+1}$ associated with the object states. Once a particular observation is received, the corresponding successor belief is obtained through the belief-update mapping
% \begin{equation}
% b_{k+1} = \mathcal{U} \left(b_k, a_k, z_{k+1} \right)
% \end{equation}
% where $\mathcal{U}$ propagates the Gaussian belief under action $a_k$ to the next decision epoch and incorporates $z_{k+1}$ through a Kalman measurement update. The propagation and measurement-update models are detailed in~\Cref{subsec:belief_tracking}. In the present formulation, observations provide noisy information about the spacecraft and secondary states. During planning, future observations are sampled from the observation model to generate possible successor beliefs. The planner therefore reasons over how future tracking information may change the estimated encounter geometry and uncertainty before TCA.
\subsection{Chance-Constrained Collision Avoidance}
\label{subsec:chance_constrained_ca}

The belief-MDP describes how the conjunction estimate may evolve under future actions and observations. We next define the collision-risk quantity associated with a belief and formulate maneuver selection subject to a chance constraint on the terminal collision risk.

\subsubsection{Probability of Collision}
\label{subsubsec:pc}

When evaluating collision risk, the current belief $b_k$ is propagated from decision epoch $k$ to TCA to obtain the corresponding predicted state distribution at closest approach. For the short-duration conjunctions considered in this work, collision risk is evaluated in the two-dimensional encounter plane at TCA, defined as the plane normal to the relative velocity vector of the two objects at closest approach~\cite{foster1992,alfano2005numerical}. Given a belief $b_k$ at the current decision epoch, the primary and secondary state distributions are propagated to TCA and combined to form the relative-position distribution. This distribution is then projected onto the encounter plane. We denote the resulting encounter-plane relative position between the primary and secondary objects at TCA by the random variable $Y_k \in \mathbb{R}^{2}$, which is normally distributed as
\begin{equation}
Y_k
\sim
\mathcal{N}
\left(
\mu_{Y,k},
\Sigma_{Y,k}
\right)
\end{equation}
where $\mu_{Y,k}$ and $\Sigma_{Y,k}$ are the mean and covariance of the relative position between the two objects, projected onto the encounter plane at TCA. These quantities are obtained by propagating the belief at decision epoch $k$ to TCA. Let $\rho$ denote the combined hard-body radius of the two objects. The probability of collision associated with belief $b_k$ is then
\begin{equation}
P_c(b_k)
=
\Pr
\left(
\lVert Y_k\rVert
\leq
\rho
\right)
\end{equation}
where $\lVert Y_k\rVert \leq \rho$ denotes the event that the relative position at TCA lies within the hard-body collision region~\cite{foster1992,alfano2005numerical}. Because the distribution of $Y_k$ is determined by the belief $b_k$, $P_c(b_k)$ depends on both the estimated encounter geometry and its associated uncertainty. The specific encounter-plane construction and numerical method used to evaluate $P_c(b_k)$ are described in the \hyperref[app:pc_evaluation]{Appendix}.

\subsubsection{Chance-Constrained Objective}
\label{subsubsec:chance_constraint}

At decision epoch $k$, the observations that will be received over the remainder of the planning horizon are not yet known. Different observation sequences may produce different beliefs at TCA $\tau = 0$. We denote this uncertain future belief by the random variable $B_{\tau=0}$. Once a particular sequence of observations is realized, the resulting terminal belief is denoted by $b_{\tau=0}$. A realized terminal belief violates the prescribed collision-risk criterion when
\begin{equation}
P_c\left(b_{\tau=0}\right) > \delta
\end{equation}
where $\delta$ is the maximum acceptable probability of collision. %From decision epoch $k$, whether this violation will occur is uncertain because $b_{\tau=0}$ depends on observations that have not yet been received. 
From the current decision epoch $k$, however, it is not yet known whether this event will occur because the terminal belief $b_{\tau=0}$  depends on future observations. %To evaluate this probability, the planner uses rollouts that sample possible sequences of future observations and resulting belief evolution, with each rollout producing a sampled terminal belief $(b_{\text{TCA}})$. 
The chance constraint therefore bounds the probability of the occurrence of this violation event with
\begin{equation}
\Pr\Big(
P_c(B_{\tau=0}) > \delta
\mid b_k,a_k
\Big)
<
\alpha
\end{equation}
where $\alpha$ specifies the maximum acceptable probability of violating the collision-risk criterion $\delta$ over future observations and the resulting belief evolution. The two parameters therefore operate at different levels: $\delta$ defines what constitutes an unsafe terminal belief with probability of collision, while $\alpha$ defines the maximum probability the planner may tolerate of reaching an unsafe terminal belief. In this work, we prescribe the probability of collision threshold as $\delta = 10^{-5}$, which is consistent with the range of ($P_c$) criteria reported by LEO operators and equals the median high-interest threshold in the operator survey of Alfano et al.~\cite{alfano2022operators}. The threshold is also more conservative than NASA's ($10^{-4}$) baseline mitigation criterion~\cite{nasa2023npr8079}. The planner's chance-constraint risk level is set to $\alpha=0.05$, balancing the desired confidence in constraint satisfaction with the need to retain feasible candidate actions. Because the chance constraint is imposed only on the terminal collision risk at TCA, intermediate beliefs along the planning horizon may have $P_c(b_k)>\delta$. These intermediate values do not by themselves make an action infeasible, since subsequent tracking observations may change the terminal belief and reduce the probability of violating the collision-risk criterion at TCA. Evaluating the constraint at TCA therefore allows the planner to account for future belief evolution without requiring a maneuver solely because an intermediate estimate exceeds the collision-risk threshold.  %the probability of terminating above the collision-risk threshold to below $\alpha$. Conversely, an action is considered infeasible when its probability of reaching TCA with $P_c > \delta$ exceeds $\alpha$.

When candidate actions satisfy the terminal chance constraint, the planning objective is to balance expected terminal collision risk against maneuver expenditure. For a candidate action $a_k$ selected from belief $b_k$, we define the root decision objective as
\begin{equation}
J(b_k,a_k)
=
-r_{\mathrm{pc}}
\mathds{E}
\left[
P_c(B_{\tau=0})
\mid
b_k,a_k
\right]
-
r_{\mathrm{man}}
\mathds{1}
\left[
a_k=\text{MANEUVER}
\right]
\label{eqn:rootDecision}
\end{equation}
where $r_{\text{pc}}$ weights the expected terminal collision probability and $r_{\text{man}}$ penalizes whether or not a maneuver was executed. The expectation $\mathds{E}[P_c(B_{\tau=0})\mid b_k,a_k]$ represents the average terminal collision probability over the possible terminal beliefs that may result after selecting action $a_k$ from the current belief $b_k$. This quantity differs from the violation probability used by the chance constraint: the violation probability measures how often the terminal collision probability exceeds $\delta$, whereas the expected terminal collision probability captures the average magnitude of collision risk across possible terminal outcomes.%The specific reward weights used by the planner are defined in~\Cref{sec:methodology}.

%The planner estimates $p_{\mathrm{viol}}(b_k,a_k)$ from belief-space Monte Carlo rollouts for each candidate action. The empirical feasibility test and subsequent action-selection procedure are described in~\Cref{sec:methodology}.

\section{Methodology}
\label{sec:methodology}

\begin{figure}[htbp]
    \centering
    \includegraphics[width=0.8\linewidth]{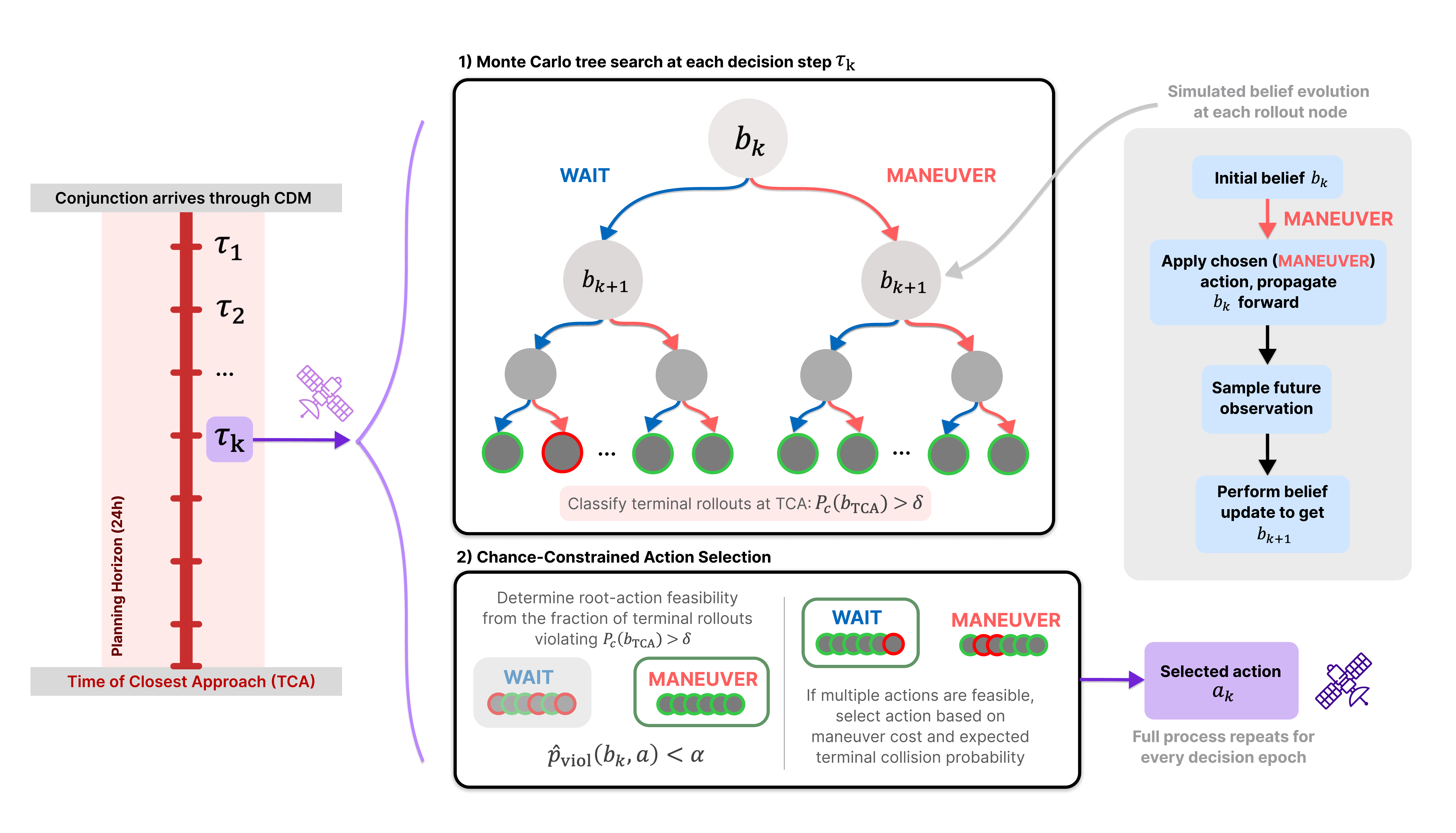}
    \caption{Receding-horizon chance-constrained belief-space planning architecture. At each decision epoch, MCTS samples possible future belief evolution under the candidate root actions and evaluates the resulting terminal collision probabilities at TCA. The fraction of terminal rollouts exceeding $\delta$ determines chance-constrained feasibility under risk level $\alpha$, after which feasible actions are ranked according to expected terminal collision risk and maneuver cost. Only the selected root action is executed before the belief is updated and planning is repeated.}
    \label{fig:planning_architecture}
\end{figure}

This section describes the planning and belief-propagation methodology used to solve the finite-horizon belief-MDP formulated in~\Cref{sec:problem_formulation}. We use a chance-constrained receding-horizon Monte Carlo tree search (MCTS)~\cite{browne2012mcts} planner to sample rollouts of possible future action-observation sequences from the current belief through TCA. The resulting terminal beliefs are used to estimate, for each candidate root action, both the probability of violating the prescribed collision-risk threshold and the expected terminal probability of collision. The former determines chance-constrained feasibility, while the latter is used together with maneuver cost to rank feasible actions. The overall planning architecture is illustrated in
\Cref{fig:planning_architecture}.

%We first describe how beliefs are propagated and updated as new tracking observations are received, followed by the evaluation of collision probability from the resulting belief. We then describe the belief-space MCTS procedure used to sample possible future belief evolution and the chance-constrained root-action selection rule used to determine the action executed at each decision epoch. Finally, we describe the receding-horizon execution procedure through which the belief is updated and the planning process is repeated as new tracking information becomes available.

% At each decision epoch, MCTS samples possible future belief evolution through TCA for each candidate root action. The resulting terminal outcomes are used to evaluate chance-constrained feasibility and select the action executed at the current epoch.

\subsection{Belief Propagation and Measurement Update}
\label{subsec:belief_tracking}

Each belief-space transition propagates the primary and secondary Gaussian beliefs to the next decision epoch and incorporates the available tracking information. For each object $j\in\{s,d\}$, the predicted mean and covariance are propagated according to
\begin{equation}
\mu_{k+1}^{j,-}
=
f^j\left(\mu_k^j,a_k,\Delta t_k\right),
\qquad
\Sigma_{k+1}^{j,-}
=
\Phi_k^j
\Sigma_k^j
\left(\Phi_k^j\right)^{\top}
\end{equation}
where $f^j$ denotes the nonlinear orbital propagation model for object $j$, and $\Phi_k^j$ is the corresponding state-transition matrix over the interval between successive decision epochs. For the primary spacecraft, $f^s$ applies the fixed prograde impulse when $a_k=\text{MANEUVER}$ before propagating the state. The secondary remains nonmaneuvering. At each tracking opportunity, object $j$ is observed through the measurement model
\begin{equation}
z_{k+1}^j
=
h\left(x_{k+1}^j\right)
+
\nu_{k+1}^j,
\qquad
\nu_{k+1}^j
\sim
\mathcal{N}\left(0,R^j\right)
\end{equation}
where $R^j$ is the measurement covariance. Observations are modeled as direct measurements of the full Cartesian state in ECI coordinates, such that
\begin{equation}
h\left(x^j\right)=x^j,
\qquad
H^j
=
\frac{\partial h}{\partial x^j}
=
I_6
\end{equation}
and the measurement update therefore follows a simplified version of the extended Kalman filter equations
\begin{align}
K_{k+1}^j & = \Sigma_{k+1}^{j,-} \left[ \Sigma_{k+1}^{j,-} + R^j \right]^{-1} \\[1ex]
\mu_{k+1}^j & = \mu_{k+1}^{j,-} + K_{k+1}^j \left( z_{k+1}^j - \mu_{k+1}^{j,-} \right) \\[1ex]
\Sigma_{k+1}^j & = \left( I_6-K_{k+1}^j \right) \Sigma_{k+1}^{j,-}
\end{align}
where the simplification follows from $H^j=I_6$. The measurement model is linear, while the belief propagation follows the nonlinear orbital dynamics. During planning, future measurements are not known and are therefore sampled from the observation model. Each sampled measurement produces a different posterior belief, providing the observation-dependent branching used by the belief-space MCTS planner. The measurement-uncertainty models used to define $R^j$, along with the tracking schedules considered in the experiments, are described in~\Cref{subsec:tracking_config}.

Each terminal belief is propagated to TCA to evaluate $P_c(b_k)$, which is used for chance-constrained feasibility and root-action ranking. The encounter-plane construction and Elrod evaluation are described in the \hyperref[app:pc_evaluation]{Appendix}.

\subsection{Belief-Space Monte Carlo Tree Search}
\label{subsec:belief_mcts}

The distribution of possible terminal beliefs $B_{\tau=0}$ cannot be enumerated directly because future tracking observations may produce different posterior beliefs over the remaining planning horizon. We therefore use Monte Carlo tree search to sample possible future action-observation sequences from the current belief through TCA. Starting from $b_k$, each MCTS rollout generates one possible sequence of future beliefs and terminates with a realization $b_{\tau=0}$. Repeated rollouts therefore provide additional samples of the possible terminal belief evolution used to evaluate the chance-constrained decision problem. 

To guide tree-search exploration, we construct an MCTS reward inspired by the maneuver-cost and terminal collision-risk components of the root decision objective $J$ in~\Cref{eqn:rootDecision}. Within the tree, each rollout accumulates a fixed penalty $r_{\mathrm{man}}$ for each maneuver executed and, at TCA, a terminal collision-risk penalty $r_{\mathrm{pc}}P_c(b_{\tau=0})$. Rollouts terminating above the collision-risk threshold additionally incur a large fixed penalty $r_{\mathrm{viol}}$. The sample-average return following action $a$ from belief $b$ defines the MCTS search value $Q(b,a)$. This value is then used to guide tree traversal according to the upper-confidence-bound criterion~\cite{kocsis2006}
\begin{equation}
a_{\mathrm{sel}}
=
\arg\max_{a\in\mathcal{A}}
\left[
Q(b,a)
+
c
\sqrt{
\frac{\ln N(b)}
{N(b,a)}
}
\right]
\end{equation}
where $N(b)$ is the number of visits to belief node $b$, $N(b,a)$ is the number of times action $a$ has been selected from that node, and the constant $c$ controls the exploration--exploitation tradeoff. Because future observations are continuous, progressive widening~\cite{sunberg2018} limits the number of observation-dependent successor beliefs expanded from each action. We highlight that the search value $Q(b,a)$ is used only to guide sampling and exploration within the MCTS tree, whereas the terminal rollout outcomes are used to evaluate chance-constrained feasibility and final selection of feasible root actions according to $J(b_k,a_k)$, as described in the following section. As an ablation, we also consider direct root action selection using $Q(b_k,a)$ rather than $J$ which we explore further in~\Cref{sec:baselines}.

The MCTS planner is executed in a receding-horizon manner. At each decision epoch, it evaluates possible belief evolution through TCA and selects the root action $a_k^\star$ using the chance-constrained procedure in~\Cref{subsec:root_chance_constraint}. Only this root action is executed. After propagating to the next decision epoch and incorporating the new tracking observation, the posterior belief $b_{k+1}$ becomes the root of a new MCTS search over the shortened horizon. This process repeats until TCA, allowing the planner to revise whether to wait or maneuver as new information becomes available.

\subsection{Chance-Constrained Root Action Selection}
\label{subsec:root_chance_constraint}

The terminal beliefs generated by MCTS are used to evaluate the chance constraint for each candidate root action. Let $\mathcal{L}(a_k)$ denote the set of rollouts beginning with action $a_k$, and let $L(a_k)=|\mathcal{L}(a_k)|$ denote the number of corresponding terminal outcomes. The violation probability introduced in~\Cref{subsubsec:chance_constraint} is estimated from the fraction of these rollouts whose terminal collision probability exceeds $\delta$
\begin{equation}
\hat{p}_{\mathrm{viol}}(b_k,a_k)
=
\frac{1}{L(a_k)}
\sum_{\ell\in\mathcal{L}(a_k)}
\mathds{1}
\left[
P_c\left(b_{\tau=0}^{(\ell)}\right)>\delta
\right]
\end{equation}
where $b_{\tau=0}^{(\ell)}$ is the terminal belief produced by rollout $\ell$. A candidate action is considered feasible when
\begin{equation}
\hat{p}_{\mathrm{viol}}(b_k,a_k)<\alpha
\end{equation}
under the given chance constraint. The same terminal outcomes are used to estimate the expected terminal collision probability appearing in the root decision objective from~\Cref{eqn:rootDecision}
\begin{equation}
\widehat{\mathds{E}}
\left[
P_c(B_{\tau=0})
\mid b_k,a_k
\right]
=
\frac{1}{L(a_k)}
\sum_{\ell\in\mathcal{L}(a_k)}
P_c\left(b_{\tau=0}^{(\ell)}\right)
\end{equation}
where these two estimates serve different purposes: $\hat{p}_{\mathrm{viol}}$ determines whether an action satisfies the chance constraint, while the expected terminal collision probability characterizes the average terminal risk used to rank feasible actions. Substituting this estimate into the root decision objective in~\Cref{eqn:rootDecision} gives its Monte Carlo estimate $\widehat{J}(b_k,a_k)$. The set of chance-constrained feasible actions is 
\begin{equation}
\mathcal{A}_{\mathrm{feas}}(b_k)
=
\left\{
a\in\mathcal{A}
\;\middle|\;
\hat{p}_{\mathrm{viol}}(b_k,a)<\alpha
\right\}
\end{equation}
and the action executed at decision epoch $k$ is selected according to
\begin{equation}
a_k^\star
=
\argmax_{a\in\mathcal{A}_{\mathrm{feas}}(b_k)}
\widehat{J}(b_k,a)
\end{equation}
and this construction allows for the separation of probabilistic safety from action ranking. The chance constraint first determines which candidate actions are admissible based on the sampled distribution of terminal collision risk. Afterwards, $\hat{J}$ selects among the feasible actions based on expected terminal risk and maneuver expenditure. However, if neither candidate action satisfies the chance constraint, the feasible set is empty. This may occur when uncertainty in the current belief prevents either action from satisfying the prescribed confidence level, even when acceptable outcomes may exist. In this case, the planner instead selects the action with the lowest estimated expected terminal collision probability
\begin{equation}
a_k^\star
=
\argmin_{a\in\mathcal{A}}
\widehat{\mathds{E}}
\left[
P_c(B_{\tau=0})
\mid b_k,a
\right]
\end{equation}
allowing the planner to select the least-risky action when neither available decision is predicted to satisfy the prescribed chance constraint under the current belief.

\section{Experiments}
\label{sec:experimental_setup}

This section describes the conjunction scenarios, tracking conditions, and comparison policies used to evaluate the proposed chance-constrained planner. The NASA Conjunction Assessment Risk Analysis Tools dataset~\cite{nasa_cara_analysis_tools} provides the basis for the evaluation, with a subset of encounters selected to match the scope of this work. Measurement quality and tracking cadence are varied to examine how the information available to the planner affects its decisions, yielding 96 evaluation configurations (8 conjunctions $\times$ 3 measurement-quality levels $\times$ 4 cadences). Baseline policies and evaluation metrics are then introduced to characterize maneuver timing, collision risk, and propellant expenditure.

Orbital states are propagated numerically using the Brahe astrodynamics library~\cite{eddy2026brahe}. Both objects are propagated in the Earth-centered inertial frame under a force model including a spherical-harmonic geopotential, atmospheric drag, third-body perturbations, and solar radiation pressure, with per-object ballistic parameters taken from the conjunction data message. Covariances are propagated by the linearized mapping obtained from the same numerical propagation. The same propagator is used to map each belief forward to TCA for collision-probability evaluation, and to back-propagate the CDM state and covariance from TCA to the start of the planning horizon.

\subsection{Conjunction Dataset and Scenario Selection}
\label{subsec:conjunction_dataset}

We evaluate the proposed planner using conjunctions drawn from the NASA CARA Analysis Tools dataset~\cite{nasa_cara_analysis_tools}. The dataset contains historical conjunctions involving operational spacecraft between 2020 and 2023, provided as Conjunction Data Messages. Each CDM provides the primary and secondary state estimates and associated covariances at TCA together with encounter quantities including miss distance, relative velocity, combined hard-body radius, and a reference probability of collision.

The CDM state and covariance information is used to initialize the conjunction beliefs evaluated by the planner. Because the CDMs provide state estimates at TCA, the object states and covariances are propagated backward from TCA to the beginning of the planning horizon. Forward propagation of these beliefs then produces the evolving state-estimation uncertainty considered throughout the decision process while retaining the encounter geometry and covariance structure of the original conjunction.

As defined in~\Cref{subsec:scenario_scope}, this work targets conjunctions for which the two-dimensional encounter-plane probability of collision provides an appropriate representation of collision risk. We apply a two-dimensional $P_c$ usage-validity assessment, following NASA CARA's implementation recommendations and usage boundaries~\cite{nasa_cara_analysis_tools}, and exclude conjunctions outside the resulting validity regime. We additionally focus on conjunctions involving a nonmaneuvering debris secondary, consistent with the maneuverable-primary and nonmaneuvering-secondary problem considered in this work. The resulting evaluation set contains 8 debris-secondary conjunctions near the operational maneuver-decision window. Table~\ref{tab:evaluation_cases} summarizes the selected encounters and their initial characteristics.

% \begin{table}[t]
% \centering
% \caption{CARA conjunction parameters used for experimental evaluation.}
% \label{tab:evaluation_cases}
% \begin{tabular}{lr}
% \toprule
% \textbf{Parameter} & \textbf{Values} \\
% \midrule
% Number of conjunctions & 8 \\
% Primary object class & Operational LEO payload \\
% Secondary object class & Debris \\
% Planning horizon ($t_0$) & 21.8--\qty{59.1}{\hour} \\
% CARA-reported $P_c$ & $1.1\times10^{-4}$--$1.1\times10^{-2}$ \\
% Miss distance & 21--\qty{448}{\meter} \\
% Combined HBR & 6.0--\qty{20.0}{\meter} \\
% Chance-constraint threshold ($\delta$) & $10^{-5}$ \\
% \bottomrule
% \end{tabular}
% \end{table}

\begin{table}[htbp]
\centering
\caption{NASA CARA conjunction dataset selected for experimental evaluation.}
\label{tab:evaluation_cases}
\begin{center}
\begin{tabular}{@{}llrrrr@{}}
\toprule
\textbf{Primary (NORAD)} &
\textbf{Secondary (NORAD)} &
\textbf{$t_0$ (h)} &
\textbf{$P_c^{\mathrm{CDM}}$} &
\textbf{Miss (m)} &
\textbf{HBR (m)} \\
\midrule
OCO-2 (40059)       & IRIDIUM 33 DEB (35921)    & 21.8 & $7.9\times10^{-4}$ & 448 &  6.0 \\
METOP-B (38771)     & FENGYUN 1C DEB (30802)    & 25.1 & $1.6\times10^{-3}$ & 148 & 10.0 \\
CALIPSO (29108)     & COSMOS 2251 DEB (34995)   & 26.3 & $1.8\times10^{-3}$ & 197 & 14.8 \\
NPP (37849)         & THOR ABLESTAR DEB (13512) & 26.5 & $1.1\times10^{-2}$ &  99 &  6.0 \\
NOAA 19 (33591)     & COSMOS 1275 DEB (42216)   & 27.0 & $4.5\times10^{-3}$ &  74 &  6.0 \\
NOAA 18 (28654)     & DMSP 5D-2 F12 DEB (41835) & 27.3 & $5.0\times10^{-3}$ &  21 &  6.0 \\
WORLDVIEW-3 (40115) & FENGYUN 1C DEB (30660)    & 27.7 & $1.1\times10^{-4}$ & 405 & 20.0 \\
TERRA (25994)       & CZ-4 DEB (26132)          & 59.1 & $1.2\times10^{-3}$ &  25 & 15.0 \\
\bottomrule
\multicolumn{6}{l}{
\footnotesize
8 conjunctions; operational LEO payload primaries; debris secondaries; $\delta=10^{-5}$
} \\
\end{tabular}
\end{center}
\end{table}
\subsection{Tracking and Measurement Configurations}
\label{subsec:tracking_config}

To evaluate how tracking information influences maneuver decisions, we independently vary the quality and cadence of measurements available for the secondary object. Measurement quality determines the uncertainty associated with each observation, while cadence determines how frequently new information becomes available before TCA.

Observations are represented in the filter as direct measurements of the full ECI state, with $H=I_6$. The measurement uncertainty, however, is constructed first in the secondary object's local radial--transverse--normal (RTN) frame. This allows the state-space covariance to retain the range-versus-cross-range structure of the underlying radar measurements, with tighter uncertainty in the range direction and larger uncertainty in the two cross-range directions. Because the belief state is maintained in ECI, the resulting RTN covariance is rotated into ECI before being supplied to the extended Kalman filter. The three measurement-quality levels in~\Cref{tab:sensor_quality} are derived from calibrated Space Surveillance Network (SSN) radar performance reported by Vallado~\cite{vallado2013} and accessed through \texttt{brahe}~1.7.0~\cite{eddy2026brahe}. The best, median, and worst configurations correspond to the 10th, 50th, and 90th percentiles of the radar noise distribution, respectively, and represent favorable, typical, and degraded tracking conditions rather than individual sensors.

For each quality level, range uncertainty defines the radial position uncertainty, while angular uncertainty is mapped to a representative cross-range uncertainty according to
\begin{equation}
\sigma_{\mathrm{cross}}
=
\sigma_{\mathrm{ang}}d
\end{equation}
using a representative sensor-target slant range of $d=\SI{1200}{km}$, approximately corresponding to a $20^\circ$ elevation angle for a target near \SI{550}{km} altitude. The measurement covariance is then constructed as
\begin{equation}
R_{\mathrm{RTN}}
=
\operatorname{diag}\!\left(
\sigma_{\mathrm{range}}^2,
\sigma_{\mathrm{cross}}^2,
\sigma_{\mathrm{cross}}^2,
\sigma_v^2,
\sigma_v^2,
\sigma_v^2
\right)
\end{equation}
where $\sigma_v=\SI{0.1}{m/s}$ and the radar range standard deviation $\sigma_{\mathrm{range}}$ is used to define the radial position uncertainty. The covariance is rotated into ECI using the secondary object's reference state at the CDM epoch and held fixed for subsequent measurement updates.

For each measurement-quality level, we additionally vary the secondary tracking cadence over $2,4,8,24$~h. These cadences represent idealized tracking schedules in which secondary measurements become available at prescribed intervals rather than through explicit sensor visibility or tasking models. The scheduled measurement times define the planner's decision epochs, such that each new secondary measurement triggers a belief update followed by a new maneuver decision. Shorter cadences therefore provide more frequent opportunities to update the conjunction belief and reconsider whether to maneuver, while longer cadences require the planner to wait longer between decisions. The primary spacecraft is assumed to have access to its own GPS-grade navigation solution (\SI{10}{m}, $1\sigma$)~\cite{hauschild2021}, which is updated at each of these decision epochs. Together, the measurement-quality and cadence sweeps allow us to evaluate when additional tracking information is sufficiently informative to justify delaying an avoidance maneuver.

\begin{table}[htbp]
\centering
\caption{Secondary-object measurement-quality configurations derived from representative SSN radars~\cite{vallado2013}}
\label{tab:sensor_quality}
\begin{center}
\begin{tabular}{@{}lrrrrr@{}}
\toprule
& & \multicolumn{2}{c}{\textbf{SSN radar noise}}
& \multicolumn{2}{c}{\textbf{Mapped state $\sigma$ (RTN)}} \\
\cmidrule(lr){3-4} \cmidrule(lr){5-6}
\textbf{Quality} & \textbf{Percentile} &
\textbf{Range (m)} & \textbf{Mean az/el (deg)} &
\textbf{Radial (m)} & \textbf{Transverse, normal (m)} \\
\midrule
Best   & 10th & 26.0  & 0.0115 & 26.0  & 241 \\
Median & 50th & 50.0  & 0.0224 & 50.0  & 469 \\
Worst  & 90th & 140.3 & 0.0477 & 140.3 & 999 \\
\bottomrule
\end{tabular}
\end{center}
\end{table}

\subsection{Baseline Policies and Ablations}
\label{subsec:baselines}

We compare the proposed chance-constrained MCTS planner against baseline decision policies designed to isolate the benefit of reasoning over future belief evolution when determining maneuver timing. Each baseline is evaluated on the same conjunctions, with the same measurement schedule, the same observation noise realizations, and the same belief update as the planner, so that any difference in outcome is attributable to the decision rule alone.

The \textit{rule-based} family holds choosing to maneuver until the time to TCA falls below a threshold $\tau_{\text{act}}$, at which point it maneuvers only if the current belief collision probability exceeds $\delta$; it then executes a single burn. This represents the common operational pattern of screening against a risk threshold on a fixed schedule, and reads $P_c$ from the same beliefs available to the chance-constrained planner. We consider decision times of 28, 12, 6, and 3~h before TCA, spanning early intervention through increasingly delayed responses. Sweeping $\tau_{\text{act}}$ traces the trade-off between acting early on poor information and deferring past the window in which a maneuver remains effective.

The \textit{optimistic greedy offline plan} represents a non-adaptive policy that commits to a maneuver strategy under favorable assumptions about future tracking. At the initial decision epoch, the policy propagates the WAIT trajectory to TCA while applying measurement updates according to the prescribed tracking schedule of 2, 4, 8, or 24 h. Each future measurement is set equal to the dynamics-predicted state, modeling an ideal measurement with no noise, so it contracts the covariance without updating the predicted trajectory. This represents planning under the optimistic, idealized assumption that the dynamics prediction is perfectly accurate, and future measurements will perfectly align, simply reducing uncertainty as TCA approaches. The policy then commits to a single decision under this optimistic forecast: if waiting is predicted to achieve $P_c < \delta$ at TCA, it commits to WAIT for the entire episode; otherwise it maneuvers immediately. This policy decision is made \textit{a priori} and does not change during evaluation. % under different simulated measurements.

This baseline is intentionally optimistic about the information that future tracking will provide. It assumes that subsequent measurements will reduce uncertainty without shifting the trajectory prediction (mean state). During execution, however, measurements are drawn from the stochastic observation model and may not continue to confirm the initial trajectory prediction. A plan that appears safe under the optimistic forecast may therefore fail to achieve the anticipated risk reduction when the realized measurements are less favorable. The baseline consequently tests the danger of committing too strongly to an expected improvement in state knowledge. By pushing deferral as far as the favorable measurement forecast permits, it provides a deliberately aggressive reference for how many maneuvers could be avoided if future tracking perfectly supported a given prediction. Its failures illustrate the corresponding cost of overconfidence in a particular, potentially incorrect, predicted future evolution of a conjunction: a plan judged sufficiently mitigated in advance may no longer satisfy the collision-risk threshold once the actual measurements are incorporated.

Finally, as an ablation, we apply the \textit{soft} $P_c$ \textit{penalty} of the chance-constrained action-selection rule, where we evaluate the same MCTS planner selecting the root action by $\arg\max_a Q(b_k,a)$. Here the collision constraint enters only softly, as the penalty $r_{\mathrm{viol}}$ embedded in the backed-up value, rather than as an explicitly estimated terminal violation probability enforced as a feasibility requirement. Comparing the two isolates the effect of separating chance-constrained feasibility from the search value used to guide tree exploration. Parameters held fixed across the test sweep are summarized in~\Cref{tab:planner_config}. All policies use the same fixed prograde maneuver magnitude and collision-risk thresholds.

\begin{table}[htbp]
\centering
\caption{Fixed planner configurations.}
\label{tab:planner_config}
\begin{center}
\begin{tabular}{@{}lr@{}}
\toprule
\textbf{Parameter} & \textbf{Value} \\
\midrule
Maneuver magnitude ($\Delta v$) & \SI{0.1}{\meter\per\second} \\
Collision-risk threshold ($\delta$) & $10^{-5}$ \\
Chance-constraint level ($\alpha$) & 0.05 \\
Collision-risk weight ($r_{\mathrm{pc}}$) & $10^{6}$ \\
Terminal violation penalty ($r_{\mathrm{viol}}$) & $10^{4}$ \\
Maneuver cost ($r_{\mathrm{man}}$) & 10 \\
MCTS exploration constant ($c$) & 10 \\
MCTS rollouts per decision & 100 \\
Full Episode Seed Runs & 5 \\
\bottomrule
\end{tabular}
\end{center}
\end{table}

\section{Results}

The results evaluate three aspects of the proposed planner: its ability to safely defer maneuvers relative to the baseline policies, the sensitivity of this behavior to tracking quality, and the mechanism by which future information influences the maneuver decision. Results are aggregated across the CARA conjunctions, measurement realizations, and tracking configurations described in~\Cref{sec:experimental_setup}.

\subsection{Safe Resolution Without Maneuver}
\label{sec:baselines}

\begin{figure}[b!]
    \centering
    \includegraphics[width=\linewidth]{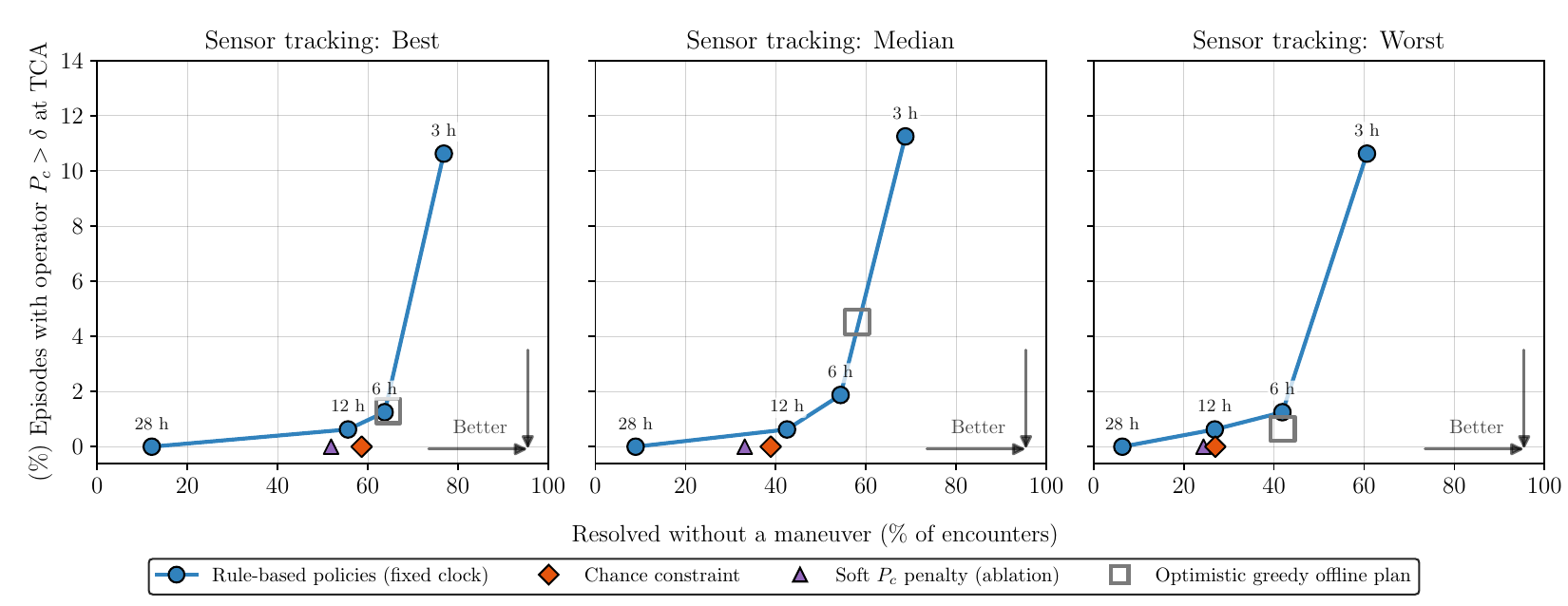}
    \caption{Encounters resolved without a maneuver and $P_c$ threshold violations at TCA for the proposed planner, its soft-penalty ablation, and comparison policies, separated by secondary measurement quality. Rule-based policies
    %, hold until a fixed time before TCA, then maneuver if the believed $P_c$ exceeds $\delta$, 
    resolve more encounters without maneuvering as $\tau_\text{act}$ approaches TCA, but more frequently violate the chance constraint, violating 10.6–11.2\% of episodes at the 3 h rule. The optimistic greedy offline plan commits to a single up-front decision under a forecast in which future measurements contract the covariance without shifting the state estimate; because that forecast overstates what tracking delivers, it still leaves 0.6–4.5\% of episodes in violation once measurements are realized. The chance-constrained planner and its soft-penalty ablation both hold violations at zero, with the chance constraint resolving more encounters without a maneuver.}
    \label{fig:baselines}
\end{figure}

We first examine how each policy balances resolving encounters without a maneuver against maintaining the prescribed collision-risk criterion at TCA. The results are shown in~\Cref{fig:baselines}.

The rule-based policies show how this balance changes with decision timing. The $28~\mathrm{h}$ rule is the most conservative, producing no observed terminal violations but resolving relatively few encounters without a maneuver. As $\tau_{\text{act}}$ moves closer to TCA, the policy benefits from additional tracking information and therefore classifies more encounters as safe to leave unmitigated. However, the decision is fixed at $\tau_{\text{act}}$ and is not revised as later measurements arrive. Those measurements can subsequently shift the estimated encounter geometry and collision risk, causing some encounters that appeared safe at the decision time to terminate above $\delta$. Thus, later decision times increase the number of scenarios resolved without a maneuver, but at the cost of higher terminal violation rates. For the $3~\mathrm{h}$ rule, approximately $61$--$77\%$ of encounters are resolved without maneuvering, with  $10.6$--$11.2\%$ of all episodes terminating above $\delta$.

The optimistic greedy offline plan approaches the same decision from a different perspective. Rather than deferring the decision to a prescribed time, it commits at the initial epoch using an optimistic forecast of future tracking. Under this policy, $41$--$64\%$ of encounters are ultimately resolved without a maneuver. When the anticipated measurements are subsequently realized, however, $0.6$--$4.5\%$ of episodes terminate above $\delta$. This result illustrates the consequence of relying too strongly on anticipated improvements in state knowledge. Future measurements can reduce uncertainty, but their realized innovations can also change the estimated encounter geometry and resulting collision risk.

The two belief-space planners occupy a different region of the comparison. Both maintain zero observed terminal violations across all three sensor-quality configurations, while still allowing a substantial fraction of encounters to reach TCA and resolve the conjunction without a maneuver. Within this zero-violation region, the chance-constrained planner resolves more encounters without maneuvering than the soft-$P_c$ penalty ablation. The chance constraint therefore improves the use of available tracking information without obtaining that improvement through an increase in observed terminal risk.

These results also clarify why the fraction of scenarios resolved without a maneuver should not be maximized in isolation. The $3~\mathrm{h}$ rule-based policy resolves the largest share of encounters without maneuvering of any policy considered---$76.9\%$ under the best tracking, against $58.6\%$ for the chance-constrained planner. However, this baseline does so while leaving $10.6\%$ of episodes above $\delta$ at TCA, while the chance-constrained planners leave none. Some conjunctions remain above the prescribed risk criterion despite additional tracking and appropriately require a maneuver; a policy that avoids maneuvering in those cases has not resolved them, it has merely declined to act. The desired outcome is therefore not $100\%$ no-maneuver resolution, but to identify which encounters can safely continue without intervention while retaining the ability to mitigate those that cannot.~\Cref{fig:baselines} shows that the chance-constrained planner moves this balance toward greater no-maneuver resolution while satisfying the terminal chance constraint.

\subsection{Effect of Tracking Quality and Frequency}
\label{sec:tracking_results}

\begin{figure}[b!]
    \centering
    \includegraphics[width=0.9\linewidth]{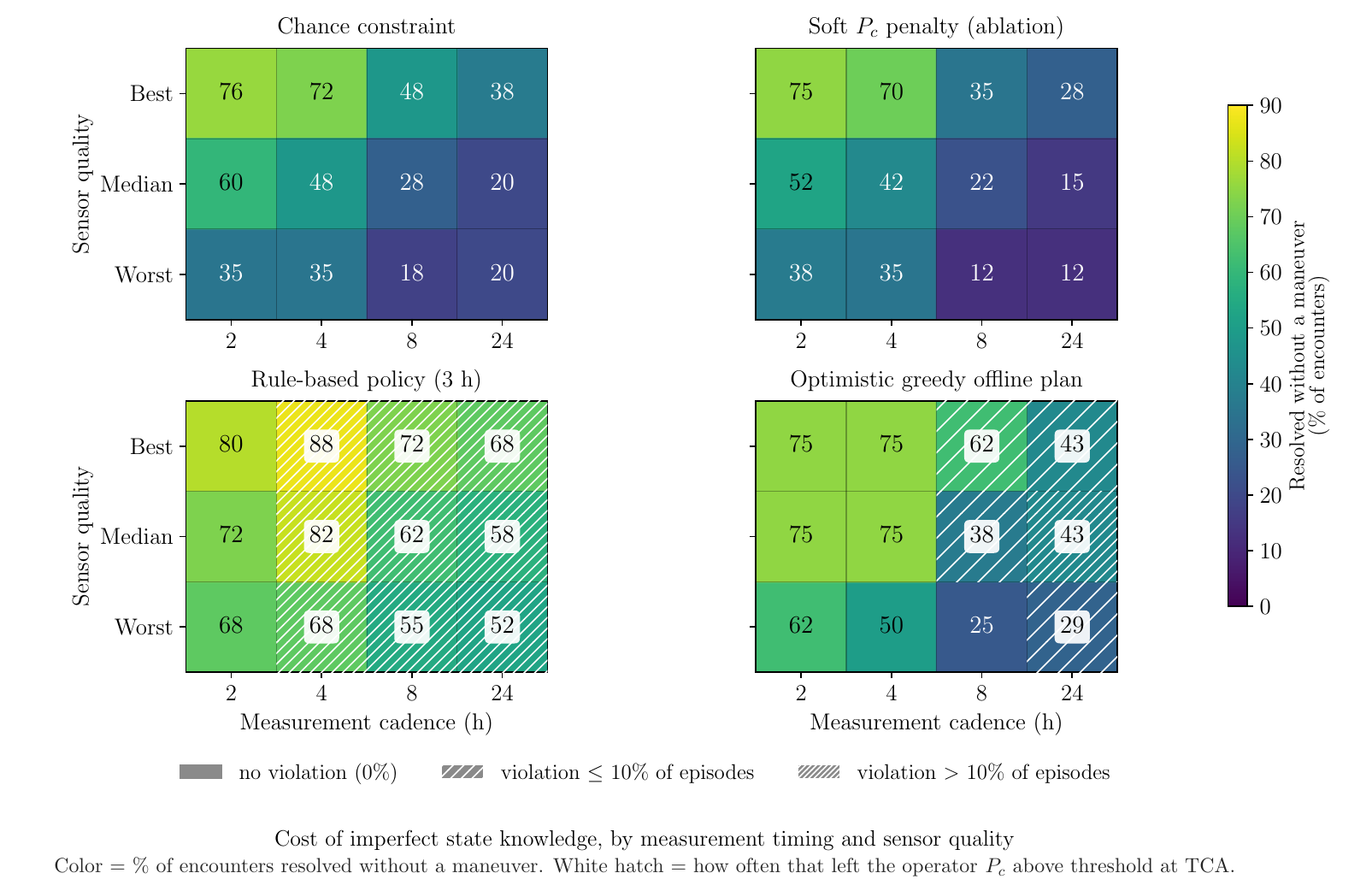}
    \caption{Encounters resolved without a maneuver across secondary measurement cadence and sensor quality. Cell color gives the percentage resolved without maneuvering, and white hatching indicates terminal $P_c>\delta$ violations. Both belief-space planners maintain zero observed violations, while the $3~\mathrm{h}$ rule-based policy and optimistic greedy offline plan reach violation rates of up to $17.5\%$ and $14.3\%$, respectively. The chance-constrained planner outperforms the soft-$P_c$ penalty ablation in eleven of twelve settings, with the largest gains under sparse measurements.}
    \label{fig:heatmap}
\end{figure}

The value of waiting depends strongly on both the quality of future measurements and how frequently they become available.~\Cref{fig:heatmap} compares the fraction of encounters resolved without a maneuver across sensor quality and measurement cadence. Because avoiding a maneuver is beneficial only when the resulting encounter remains safe, the figure reports this outcome together with the frequency of terminal collision-risk threshold violations.

For the chance-constrained planner, accurate and frequent measurements create substantial opportunity to defer intervention safely. Under the best sensor-quality configuration, $76\%$ of encounters are resolved without a maneuver at a $2~\mathrm{h}$ cadence and $72\%$ at a $4~\mathrm{h}$ cadence. This fraction decreases to $48\%$ and $38\%$ as the measurement cadence increases to $8$ and $24~\mathrm{h}$, respectively. The same trend appears as measurement quality degrades. At median sensor quality, the fraction resolved without maneuvering falls from $60\%$ at a $2~\mathrm{h}$ cadence to $20\%$ at $24~\mathrm{h}$. Under the worst sensor quality, no-maneuver resolution remains at or below $35\%$ across all cadences. Importantly, all conjunctions resolved without maneuvering have satisfied the terminal chance constraint. The interaction between measurement quality and cadence further shows that measurement frequency is most valuable when each observation is sufficiently informative. More frequent measurements substantially increase safe deferral under the best and median sensor qualities, whereas the benefit is smaller under the worst-quality configuration; even best-quality measurements at a $24~\mathrm{h}$ cadence slightly outperform worst-quality measurements received every $2~\mathrm{h}$.

The comparison with the soft-$P_c$ penalty ablation shows that explicitly constraining risk becomes particularly important when measurements are sparse. The chance-constrained planner resolves more encounters without maneuvering across the majority of tracking scenarios, with the largest separation occurring at the $8$ and $24~\mathrm{h}$ cadences. As measurements become more frequent, the two belief-space planners converge, indicating that frequent information can partially compensate for differences in how risk is incorporated into the decision rule.

The non-adaptive policies illustrate why no-maneuver resolution cannot be interpreted independently of safety. In particular, the optimistic greedy offline plan assumes favorable future measurement outcomes and commits at the initial decision epoch to the least-conservative plan that appears safe under that forecast. This assumption allows the plan to defer aggressively in several tracking regimes, but the apparent improvement comes with terminal threshold violations of up to $14.3\%$. The $3~\mathrm{h}$ rule-based policy exhibits the same tradeoff more strongly, reaching violation rates of up to $17.5\%$. Thus, greater willingness to wait does not by itself indicate better use of future information.

Together, these results show that future tracking has substantial decision value, but that value depends on accounting for uncertainty in what future measurements will reveal. Accurate and frequent measurements allow the chance-constrained planner to avoid unnecessary maneuvers while maintaining the prescribed risk criterion. Conversely, planning as though future measurements will be sufficiently favorable can make deferral appear more attractive than it is during execution. The benefit of belief-space planning is therefore not simply that it waits for more information, but that it determines when waiting remains justified under uncertainty in that information.

\begin{figure}[h!]
    \centering
    \includegraphics[width=\linewidth]{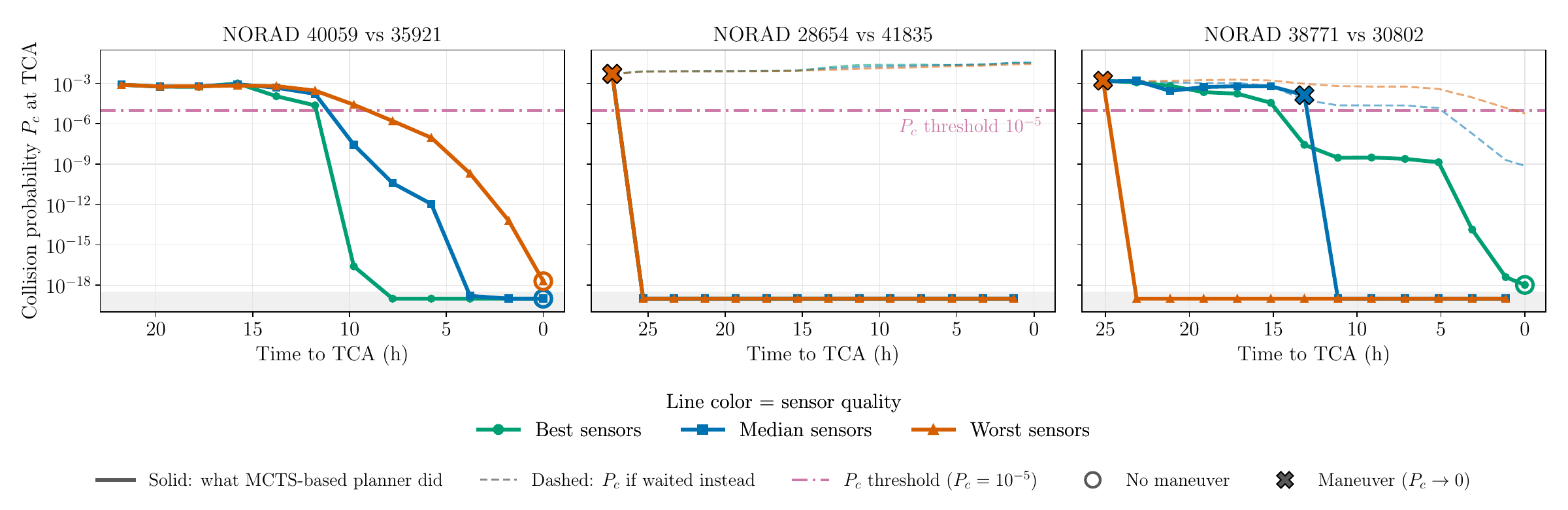}
    \caption{Belief collision probability through TCA for three representative conjunctions under best, median, and worst secondary measurement quality. Solid curves show the executed MCTS trajectory, while dashed curves show the counterfactual continued-WAIT trajectory. The examples illustrate encounters that resolve through tracking alone (left), require mitigation (center), and produce measurement-dependent maneuver decisions (right). These cases are drawn from across the sweep to illustrate the wait-and-measure behavior of the tree-based planner rather than to compare the chance-constrained and soft-$P_c$ root-selection rules.}
    \label{fig:belief}
\end{figure}

\subsection{Information-Dependent Maneuver Timing}
\label{sec:decision_behavior}

For the MCTS-based planners, the same conjunction can lead to different maneuver decisions as the quality of tracking information available before TCA changes. To examine this behavior in more detail,~\Cref{fig:belief} shows complete episodes for three representative conjunctions selected from across the sweep. The cases illustrate three distinct responses to additional tracking information: an encounter that resolves without a maneuver, one that requires mitigation in all tracking scenarios, and one for which the maneuver decision changes with measurement quality. The examples are intended to illustrate the common wait-and-measure behavior of the MCTS-based planners rather than highlight differences between the chance-constrained and soft-$P_c$ root-selection rules.

For NORAD 40059 vs.\ 35921, continued tracking reduces the estimated collision probability below $\delta$ under all three measurement-quality configurations. The planner therefore reaches TCA without maneuvering, although the threshold is crossed earlier when measurements are more accurate. In contrast, NORAD 28654 vs.\ 41835 remains hazardous under the continued-WAIT trajectory for every tracking configuration, and the planner maneuvers at the initial decision epoch. NORAD 38771 vs.\ 30802 lies between these cases. With the best tracking configuration, the planner continues to wait and ultimately reaches TCA below $\delta$ without a maneuver. With median-quality tracking, it initially waits for additional information but later executes a maneuver as the belief evolves. Under the worst tracking configuration, it maneuvers immediately. The same conjunction therefore transitions from no maneuver, to a later maneuver, to immediate intervention as measurement quality decreases.

These examples show how tracking quality changes both whether and when a maneuver is required. For some conjunctions, additional measurements consistently provide enough information to continue without intervention, while others require mitigation regardless of tracking quality. Between these cases are conjunctions for which the maneuver decision depends directly on the information obtained as the belief evolves toward TCA.

\subsection{Effect of Chance-Constrained Root Selection}
\label{sec:root_ablation}

The chance constraint and soft-$P_c$ penalty use the same MCTS search and differ only in how they select the final action. The soft penalty averages the penalized returns across rollouts, so even a single threshold violation can strongly reduce an action's value. The chance constraint instead considers how often violations occur, allowing the action to remain feasible as long as the estimated violation frequency remains below the chance constraint set by $\alpha$.

\begin{figure} [htbp]
    \centering
    \includegraphics[width=\linewidth]{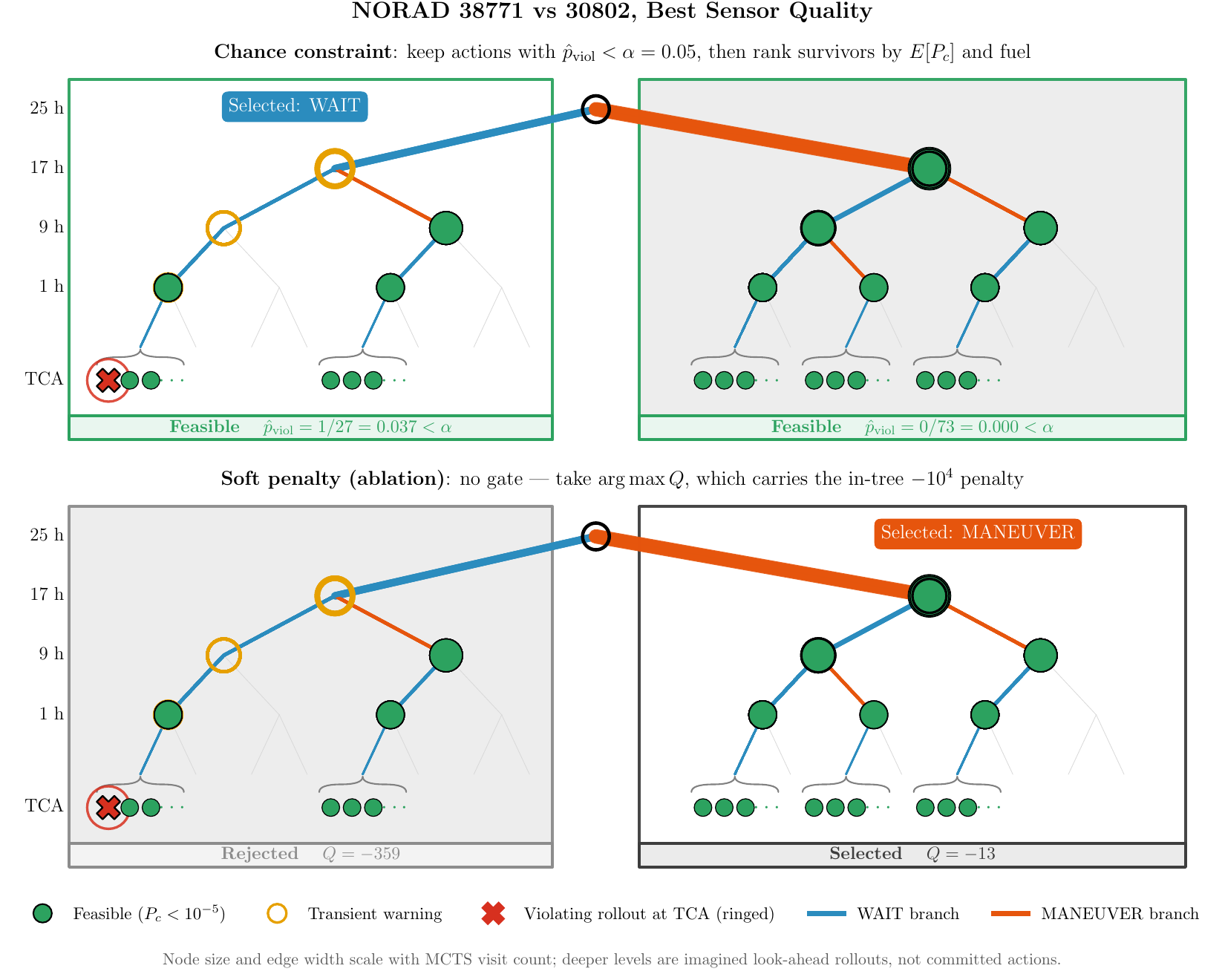}
    \caption{One MCTS root decision for NORAD 38771 vs.\ 30802 under best-quality tracking and an $8~\mathrm{h}$ measurement cadence. Both root rules read the same 325-node search tree. The chance constraint accepts WAIT because only one of 27 rollouts violates the terminal threshold, giving $\hat{p}_{\mathrm{viol}}=0.037<\alpha=0.05$, and selects the lower-cost feasible action. The soft-$P_c$ penalty instead allows this single violating rollout to dominate WAIT's mean return and selects MANEUVER. For this conjunction, \textbf{waiting is ultimately preferable}: under high-quality tracking, subsequent measurements resolve sufficient uncertainty for the encounter to reach TCA below the collision-risk threshold without a maneuver. The chance constraint preserves this option despite a single adverse rollout that would otherwise trigger unnecessary intervention.}
    \label{fig:treebest}
\end{figure}

\Cref{fig:treebest} illustrates this mechanism for a single root decision under best-quality tracking. WAIT violates the terminal threshold in one of its 27 rollouts, giving $\hat{p}_{\mathrm{viol}}=0.037<\alpha=0.05$, and therefore remains feasible. Its expected collision probability is $\mathbb{E}[P_c]=4.9\times10^{-7}$, approximately twenty times below $\delta$, so the chance-constrained rule selects WAIT as the lower-cost feasible action. Under the soft-$P_c$ penalty, the same violating rollout drives $Q(\mathrm{WAIT})=-359$ compared with $Q(\mathrm{MANEUVER})=-13$, causing the planner to maneuver.

Across the 96 paired comparisons in the sweeps examined in~\Cref{sec:baselines} and~\Cref{sec:tracking_results}, the two root rules disagree in 23 cases. In every disagreement, the chance-constrained rule selects WAIT while the soft-$P_c$ penalty selects MANEUVER; the reverse is never observed. These disagreements concentrate near the constraint boundary, where a small number of violating rollouts can substantially reduce the mean return without causing the estimated violation frequency to exceed $\alpha$. Despite selecting WAIT more often in these cases, the chance-constrained planner produces no observed increase in terminal threshold violations, with both belief-space planners maintaining zero violations across all three tracking tiers.

This behavior does not persist when the available tracking information no longer supports waiting. For the same conjunction under the worst sensor quality, WAIT has $\mathbb{E}[P_c]=1.1\times10^{-3}$, more than two orders of magnitude above $\delta$, and both rules select MANEUVER. The chance-constrained rule eliminates WAIT as infeasible, while the soft-$P_c$ rule rejects it through its mean return. Thus, the chance-constrained root rule preserves WAIT when its estimated violation frequency remains within the prescribed bound, but selects a maneuver when the available tracking information no longer supports WAIT as a feasible action.

\section{Conclusions}
\label{sec:conclusions}

Collision avoidance decisions depend not only on the collision risk estimated from the information available now, but also on the information expected to become available before TCA. For encounters with an unmaneuverable secondary, this creates a fundamental tradeoff: maneuvering early preserves mitigation authority but commits to an action using a less-informed estimate, while waiting can resolve uncertainty but leaves less time to respond. The chance-constrained belief-space formulation developed here makes this tradeoff explicit by evaluating whether future tracking is sufficiently likely to produce an acceptable terminal belief before allowing the planner to defer intervention.

The experiments show that this distinction can materially change maneuver decisions. Across the evaluated NASA CARA conjunctions and tracking conditions, the chance-constrained planner resolved approximately 40\% of episodes without a maneuver while maintaining zero terminal collision-risk violations. Rule-based policies deferred intervention more aggressively, but only by moving along a clear safety--deferral tradeoff, with threshold violation rates increasing to approximately 11\% for the latest decision policy. The belief-space planner instead reasoned explicitly over possible future tracking outcomes to determine when deferral remained admissible under the prescribed terminal risk requirement.

More importantly, the results show that the value of waiting is determined by the tracking environment. Under the most accurate and frequent secondary tracking considered, 76\% of episodes were resolved without maneuvering. As measurements became less accurate or less frequent, safe deferral decreased substantially, reaching approximately 18–20\% under the least informative measurement conditions. This transition is not simply a change in the precision of the resulting collision-probability estimate. For the same NORAD 38771 and 30802 conjunction, the planner waited through TCA under high-quality tracking but maneuvered at the first decision epoch under low-quality tracking. The physical encounter was unchanged; only the information expected to arrive before TCA differed. 

This result has a broader implication for autonomous conjunction management. Tracking performance and maneuver planning should not be treated as independent components of the response pipeline. The quality and timing of future observations determine how long an avoidance decision can safely remain open, and therefore have direct operational value in units of maneuver deferral. Conversely, degraded state knowledge carries a decision cost: when future observations cannot be expected to resolve the encounter with sufficient confidence, the planner must surrender the option to wait and commit earlier. The terminal chance constraint provides a principled boundary between these regimes by identifying when the uncertainty in future belief evolution makes continued deferral itself inadmissible.

The formulation developed here isolates this information--action tradeoff using a fixed impulsive maneuver, prescribed tracking opportunities, and short-duration conjunctions within the validity regime of the two-dimensional encounter-plane collision model. These assumptions provide a controlled setting in which the decision value of future information can be separated from continuous maneuver optimization and sensor scheduling. Extending the action space to continuous maneuver design and allowing future observations to be actively tasked would remove that separation and create a more general problem: jointly deciding when to observe, when to wait, and when and how to maneuver. In that setting, tracking resources and propellant become coupled resources for managing collision risk. The results presented here suggest that this coupling is consequential: improving knowledge of an encounter can eliminate the need to maneuver, while insufficient knowledge can itself become a reason to act.

\section*{ACKNOWLEDGMENTS}

This research was supported by the Fannie and John Hertz Foundation and the National Defense Science and Engineering Graduate (NDSEG) Fellowship Program. Specifically, this material is based upon work supported by the Air Force Office of Scientific Research under award number FA9550-25-C-B010 in the amount of currently negotiated tuition and stipend rates.

\section*{References}

\bibliographystyle{unsrt}

\begingroup
\renewcommand{\section}[2]{}%
\bibliography{references}
\endgroup

% \section*{appendix}
\appendix
\section*{Appendix}
\label{app:pc_evaluation}
\paragraph{Collision-Probability Evaluation.} For each belief encountered during planning, the probability of collision is evaluated at TCA and used both to test chance-constrained feasibility and to score the decision objective. Starting from a belief $b_k$, the mean state and covariance of each object are propagated to TCA using the dynamics and covariance propagation of~\Cref{subsec:belief_tracking}. Importantly, $\Sigma_{\tau=0}$ is obtained by propagating the \emph{accumulated} belief covariance forward from $b_k$, so that the information gained from observations taken before $t_k$ is reflected in the terminal risk; it is not reset to the initial covariance. The position covariances are combined to give the relative-position covariance
\begin{equation}
\Sigma_{\mathrm{rel}}
=
\Big(\Sigma_{\tau=0}^s\Big)^{(r)}
+
\Big(\Sigma_{\tau=0}^d\Big)^{(r)}
\end{equation}
where $\left(\Sigma_{\tau=0}^s\right)^{(r)}$ and $\left(\Sigma_{\tau=0}^d\right)^{(r)}$ are the position submatrices of the propagated primary and secondary covariances, respectively. The corresponding relative-position mean is
\begin{equation}
\mu_{\mathrm{rel}}
=
\Big(\mu_{\tau=0}^s\Big)^{(r)}
-
\Big(\mu_{\tau=0}^d\Big)^{(r)}
\end{equation}
Because the conjunctions considered here are short-duration, high relative-velocity encounters, the relative motion is approximately linear and the covariance approximately static over the encounter, permitting the standard two-dimensional reduction~\cite{foster1992,alfano2005numerical}. Let
\begin{equation}
v_{\mathrm{rel}}
=
\left(\mu_{\tau=0}^s\right)^{(v)}
-
\left(\mu_{\tau=0}^d\right)^{(v)},
\qquad
\hat{v}_{\mathrm{rel}}
=
\frac{v_{\mathrm{rel}}}{\lVert v_{\mathrm{rel}}\rVert}
\end{equation}
and let $\hat{u}_1,\hat{u}_2$ be an orthonormal basis for the encounter plane normal to $\hat{v}_{\mathrm{rel}}$. Stacking these basis vectors as rows of
\begin{equation}
T =
\begin{bmatrix}
\hat{u}_1^\top \\
\hat{u}_2^\top
\end{bmatrix}
\in \mathbb{R}^{2\times3}
\end{equation}
gives the projected relative-position mean and covariance
\begin{equation}
\mu_{Y,k} = T\mu_{\mathrm{rel}},
\qquad
\Sigma_{Y,k} = T\Sigma_{\mathrm{rel}}T^\top
\end{equation}
and the collision probability associated with belief $b_k$ is
\begin{equation}
P_c(b_k)
=
\int_{\lVert y\rVert\leq\rho}
\mathcal{N}\!\left(
y;
\mu_{Y,k},
\Sigma_{Y,k}
\right)\,dy
\end{equation}
where $\rho$ is the sum of the primary and secondary hard-body radii~\cite{foster1992,alfano2005numerical}.

We evaluate this integral using the fixed-order Gauss--Chebyshev quadrature formulation described by Elrod~\cite{elrod2019computational} implemented in NASA CARA's software toolbox~\cite{nasa_cara_analysis_tools}. In contrast to Chan's series formulation, which obtains the anisotropic case through an approximation to an isotropic formulation, the quadrature directly evaluates the one-dimensional integral associated with the general two-dimensional Gaussian encounter-plane distribution~\cite{alfano2005numerical}. This is well suited to the eccentric, potentially rotated encounter-plane covariance ellipses produced by our sensor model in~\Cref{subsec:tracking_config}. 

% \section{Submitting the paper}

% Poster and oral presenters are required to submit a technical paper prior to the Technical Paper Deadline. Please upload your technical paper in PDF format. Filename should include last name of presenter and date of upload.  Upload your paper online by logging into your abstract submission account.

\end{document}